\documentclass{article}
\usepackage{spconf,amsmath,graphicx,xcolor}
\usepackage{multirow}
\usepackage{booktabs}
\usepackage{hyperref}
\usepackage{xspace}
\usepackage{amsfonts}
\usepackage[english]{babel}

\title{UTAL-GNN: Unsupervised Temporal Action Localization using Graph Neural Networks}
\name{  Bikash Kumar Badatya$^{\dagger}$  Vipul Baghel$^{\dagger}$  Ravi Hegde }

\address{ Indian Institute of Technology Gandhinagar, Gujarat, India. 
}

\begin{document}
%
\maketitle
\begin{abstract}
Fine-grained action localization in untrimmed sports videos presents a significant challenge due to rapid and subtle motion transitions over short durations. Existing supervised and weakly supervised solutions often rely on extensive annotated datasets and high-capacity models, making them computationally intensive and less adaptable to real-world scenarios. In this work, we introduce a lightweight and unsupervised skeleton-based action localization pipeline that leverages spatio-temporal graph neural representations. Our approach pre-trains an Attention-based Spatio-Temporal Graph Convolutional Network (ASTGCN) on a pose-sequence denoising task with blockwise partitions, enabling it to learn intrinsic motion dynamics without any manual labeling. At inference, we define a novel Action Dynamics Metric (ADM), computed directly from low-dimensional ASTGCN embeddings, which detects motion boundaries by identifying inflection points in its curvature profile. Our method achieves a mean Average Precision (mAP) of \textbf{82.66\%} and average localization latency of \textbf{29.09 ms} on the DSV Diving dataset, matching state-of-the-art supervised performance while maintaining computational efficiency. Furthermore, it generalizes robustly to unseen, in-the-wild diving footage without retraining, demonstrating its practical applicability for lightweight, real-time action analysis systems in embedded or dynamic environments.
\end{abstract}

\begin{keywords}
Sports Analytics, Skeleton-based Action Localization, Graph Convolution, Representation Learning, Interpretability
\end{keywords}
\section{Introduction}
\label{sec:intro}

Sports analytics has transformed performance measurement, strategy optimization, and training methodologies~\cite{cardenas2024beyond}. Data-driven approaches enable coaches and athletes to refine techniques, analyze gameplay dynamics, and enhance decision-making~\cite{li2020graph}. Beyond performance analysis, these insights also benefit broadcasters by automating content generation and improving real-time event coverage.

Action recognition involves classifying activities in trimmed videos, while action spotting focuses on detecting precise action segments in untrimmed videos. Temporal localization determines action boundaries, either by identifying start and end timestamps or by pinpointing key transition moments. Extending this problem to spatial dimensions leads to spatio-temporal action detection, which is particularly challenging in diving due to rapid and intricate movements such as takeoffs, rotations, twists, and water entries.

Fine-grained sports action localization entails three primary tasks: determining action boundaries, classifying action categories, and interpreting motion dynamics. While prior research has primarily addressed boundary detection and classification, explainability and interpretability of motion features remain underexplored, limiting the effectiveness of existing methods. Deep learning-based approaches for this task fall into three categories: (1) fully supervised, (2) weakly supervised, and (3) unsupervised learning.

In sports, where annotated data is scarce, unsupervised learning offers advantages in scalability, adaptability, and generalization to unseen scenarios. However, the absence of labels makes fine-grained action localization in untrimmed videos significantly more challenging. To address these challenges, we propose a novel skeleton-based unsupervised learning framework for action localization, with a focus on springboard diving sport known for its precision and technical complexity. Recent computer vision and machine learning techniques enable detailed motion analysis, offering critical insights into a diver’s technique, strengths, and areas for improvement. This is crucial for optimizing training strategies and improving competitive performance.

The main contributions of this paper are the following:

\begin{itemize}
    \item We introduce an  Attention-based Spatio-Temporal Graph Convolutional Network (ASTGCN)~\cite{guo2019attention} encoder, pre-trained via blockwise pose-sequence learning on denoising task. This approach captures blockwise motion dynamics and computes a novel Action Dynamics Metric (ADM) based on the curvature of the learned embedding norms, which enables detection of motion transitions without any form of auxiliary supervision.

\footnote{This paper has been accepted at the ICIP Satellite Workshop 2025.}

    \item Our model achieves performance comparable to supervised methods on the DSV Diving dataset. Additionally, we demonstrate the generalizability of our approach in out-of-distribution, in-the-wild diving videos.
    \item We provide a graphical representation of the learned embeddings as a measure of pose dynamics and transitions. Furthermore, we provide a theoretical justification demonstrating that inflection points correspond to action transition states.
\end{itemize}

This work contributes to advancing fine-grained action transition understanding in computer vision, paving the way for more robust and interpretable models.

\section{Related Work}
\label{sec:format}
\subsection{Fully Supervised Temporal Action Localisation}

Fully supervised temporal action localization methods are categorized into one-stage, two-stage, and anchor-free pipelines based on their dependence on anchors.

One-stage methods, such as those incorporating appearance and motion fusion~\cite{yang2020revisiting}, offer efficiency but are constrained by default anchors, requiring meticulous hyperparameter tuning. Two-stage approaches generate action proposals using predefined temporal anchors, refining them via boundary regression and classification. For example,~\cite{zhao2020bottom} employs an attention-based graph convolutional module to enhance proposal relationships. In contrast, anchor-free pipelines directly predict action boundaries without predefined anchors, demonstrating robustness in handling varied action durations~\cite{liu2021multi}.

Temporal action localization further employs frame classification and proposal classification strategies. Frame classification captures spatial information per frame and models temporal dynamics through predefined rules; for instance,~\cite{islam2021hybrid} estimates action start and end probabilities. Proposal classification focuses on generating and classifying action proposals, leveraging dense enumeration strategies as in~\cite{lee2020background}.

\subsection{Weakly Supervised Temporal Action Localisation}
In weakly supervised temporal action localization, learning is constrained to video-level classification labels, while accurate localization requires frame-level predictions.

Pre-classification methods apply temporal convolutions to classify individual frames, aggregating scores via weighted sum~\cite{shi2020weakly} or top-k mean~\cite{min2020adversarial} to derive video-level labels. Post-classification follows a two-step approach: assessing frame relevance to video-level classification via attention weights, then aggregating video features into a single vector for classification. 

\subsection{Unsupervised Temporal Action Localisation}

Unsupervised Temporal Action Localization (UTAL)~\cite{zhang2022automatic} enables action localization in unlabeled videos, requiring only the total count of action categories, thereby reducing annotation costs. Fang et al.~\cite{fang2024bid} proposed an unsupervised skeleton-based pretraining framework that segments motion sequences into meaningful components, extracting action features without annotations. To address UTAL challenges, Tang et al.~\cite{tang2023unsupervised} introduced FEEL, a self-paced incremental learning model, incorporating a clustering confidence improvement module with feature-robust Jaccard distance for refined video clustering and enhanced label prediction. These approaches collectively advance temporal action localization by improving accuracy, efficiency, and generalization across diverse action instances. However, these methods rely heavily on clustering and self-paced learning, which lack explicit motion dynamics modeling and struggle with fine-grained transitions in sports actions.

\subsection{Sports Domain}
Complex player interactions make sports analysis of dynamic action scenarios difficult. Seweryn et al.~\cite{seweryn2023survey} explored soccer action localisation using multimodal techniques, including video, audio, and different data. A fine-grained diving dataset and procedure-aware cross-attention-based temporal segmentation were developed by Xu et al.~\cite{xu2022finediving} for detailed action recognition. These approaches can't handle different sports and unstructured motion data since they depend on human annotations and domain-specific heuristics. They also prioritise categorisation and coarse action segmentation over sub-action transitions, which are essential for comprehending fine-grained sports motions.

\section{Dataset Pre-processing}
\label{sec:pagestyle}
We use the DSV-diving dataset~\cite{murthy2023divenet} to train and validate the proposed pipeline. The data set consists of various dive actions performed at four different heights of the spring: 3m, 5m, 7.5m, 10 meters. Each diving action lasts for 2 to 5 seconds within a single video sequence recorded at 60 fps. The annotations in the dataset consist of 2D coordinates of 16 key body joints in MPII format for each diving clip along with 5 action demarcations, denoted as 'start', 'm1', 'm2', 'm3', and 'end'. For training and inferencing purpose, each pose sequence is partitioned into sub-pose sequences with a rolling window of size W in shuffled and serial manner respectively.  

\section{Methodology}
\label{sec:typestyle}

Previous studies like ~\cite{chen2020curvature} proposed geometric curvature-based  encoding for supervised action recognition. The authors encoded the action pattern
into curvatures on the global timescale. Inspired by this, we propose a skeleton-based unsupervised pipeline for fine-grained sports action detection in untrimmed videos. Our approach leverages an Attention-Based Spatial-Temporal Graph Convolutional Network (ASTGCN)~\cite{guo2019attention} to extract spatio-temporal graph embeddings from pose sequences.
\begin{figure}[ht]
    \centering
    \includegraphics[width=0.48\textwidth, height=0.55\textwidth]{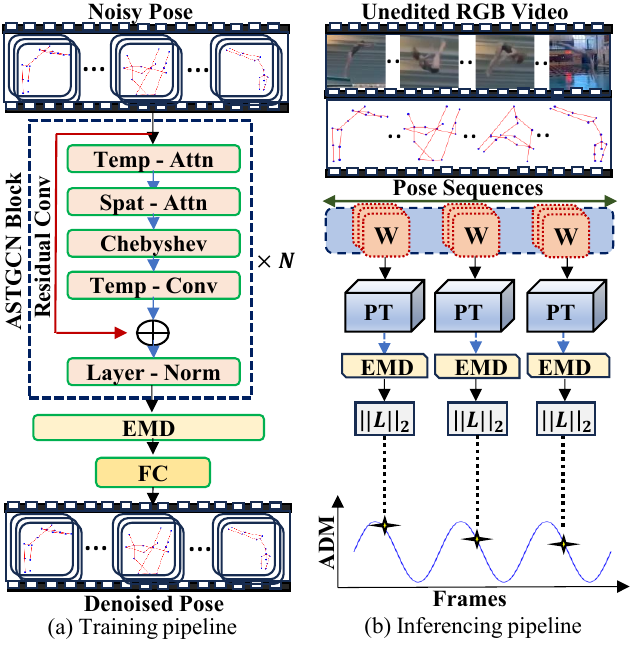}
\caption{
(a) \textbf{Training:} Noisy pose sub-sequences are denoised via ASTGCN~\cite{guo2019attention} blocks to generate spatio-temporal embeddings (EMD), followed by a Fully Connected (FC) layer. 
(b) \textbf{Inference:} Blockwise clean pose with window size $W$ are passed through the pre-trained model (PT) to extract EMD. The $\lVert L \rVert_2$ norm forms the Action Dynamics Metric (ADM), whose curvature identifies inflection points as action transitions.
}
\label{fig:fig1}
\end{figure}

\subsection{Training}
As shown in Figure~\ref{fig:fig1}(a), the proposed pre-training denoising task operates on blockwise partitioned pose sequences instead of full video sequences. Each sequence is segmented into overlapping sub-sequences of window size W using a rolling window (stride = 1), followed by shuffling. Each sub-sequence is treated as a spatio-temporal graph with 2D coordinates as node attributes. Noisy pose sub-sequences, generated by adding Gaussian noise, are processed through N ASTGCN layers that leverage attention-based graph convolutions to encode spatial and temporal dependencies. The output of the ASTGCN blocks is an embedding (EMD) of the input pose sequence. This embedding is then passed through a Fully Connected (FC) layer to reconstruct the denoised pose, optimized via Mean Square Error (MSE) loss.

\subsection{Inferencing}
During inference, as shown in Figure~\ref{fig:fig1}(b), pose sequences are sequentially partitioned into sub-sequences using a rolling window of size $W$. Refined $D$-dimensional spatio-temporal embeddings are extracted via the pre-trained model (PT). To analyze motion variations, the Euclidean norm of these embeddings is computed over spatial and temporal axes, yielding a single-point representation termed as the \emph{Action Dynamics Metric} (ADM). ADM captures pose dynamics evolution, and curvature changes in its sequence identify inflection points corresponding to significant action transitions. Intuitively, curvature quantifies how sharply the ADM signal bends over time. Smooth and consistent motions yield a low curvature profile, while abrupt transitions in action—such as takeoff, twist, or entry—cause sharp changes in ADM values, resulting in noticeable curvature spikes. These spikes indicate temporal boundaries between sub-actions, allowing us to detect key transition phases in an unsupervised manner.

\begin{figure*}[t!]
    \centering
    \includegraphics[width=1.0\textwidth]{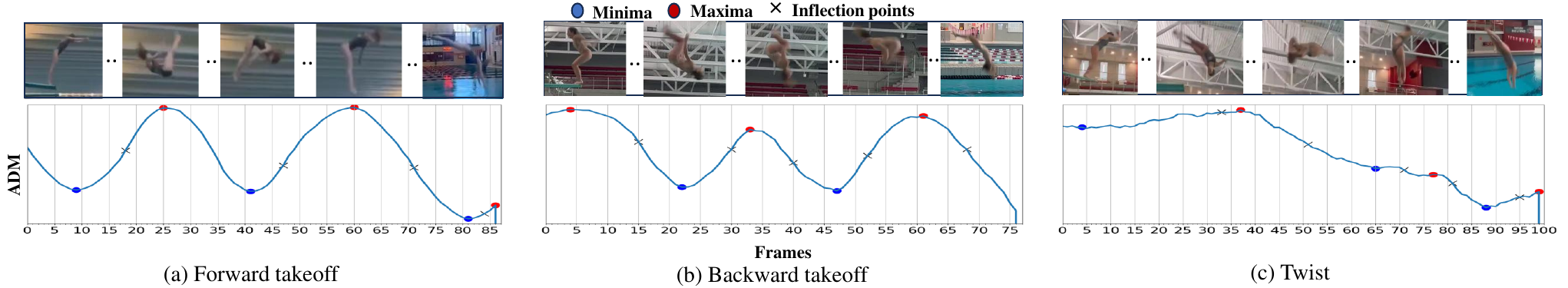}
    \caption{Interpretation of ADM curve and significant action transitions detection for three distinct diving classes.}
    \label{fig:fig2}
\end{figure*}

\section{Theoretical Justification: Inflection Points as Motion Transitions}

\subsection{Problem Setup}
Let the input pose sequence be \( X \in \mathbb{R}^{B \times F \times J \times C} \), where \( B \), \( F \), \( J \), and \( C \) denote batch size, time steps, joints, and feature dimension, respectively. Define a human pose graph \( G = (V, E) \), with adjacency matrix \( A \in \mathbb{R}^{J \times J} \). The ASTGCN embedding is denoted by:
\[
Z = f_{\text{ASTGCN}}(X, A) \in \mathbb{R}^{B \times F \times J \times C}
\]
We define the embedding norm sequence:
\[
S_b = \| Z_b \|_2 = \sqrt{\sum_{f,j,c} Z_{b, f, j, c}^2}, \quad b = 1, \dots, B
\]
and its discrete second derivative:
\[
\Delta^2 S_b = S_{b+1} - 2S_b + S_{b-1}
\]
A candidate transition point is characterized by \( \Delta^2 S_b \approx 0 \) and a sign change in its discrete curvature: \( \Delta^2 S_{b-1} \cdot \Delta^2 S_{b+1} < 0 \), indicating a local extremum or inflection in the ADM signal.

\subsection{Intuition Behind Transition Detection}
Suppose two motion phases \( A_1 \) and \( A_2 \) are separated at frame \( b_T \). The embedding function changes across this boundary:
\[
Z_b = \begin{cases}
    f_1(X_b, A), & b \leq b_T \\
    f_2(X_b, A), & b > b_T
\end{cases}
\]
Although \( S_b \) is a smooth function within each phase, its curvature may shift abruptly at the transition:
\[
\lim_{b \to b_T^-} \frac{d^2 S}{db^2} \neq \lim_{b \to b_T^+} \frac{d^2 S}{db^2}
\]
Given:
\[
S_{b+1} - S_b = \| f_{\text{ASTGCN}}(X_{b+1}, A) \|_2 - \| f_{\text{ASTGCN}}(X_b, A) \|_2
\]
a sudden shift in the embedding dynamics implies:
\[
\frac{d^2}{db^2} \| f_{\text{ASTGCN}}(X_b, A) \|_2 \not\approx 0
\]
Therefore, by detecting curvature changes in \( S_b \) (i.e., inflection points), we can heuristically identify temporal boundaries between motion phases.

\section{Results and Discussion}
\label{sec:majhead}
We evaluate our approach on the DSV diving dataset~\cite{murthy2023divenet}, which provides ground-truth 2D poses and temporal demarcations with separate training and testing partitions. During training, we use a rolling window size of W = 7 and Gaussian noise standard deviation of $\sigma$ = 0.1 to generate noisy input sub-pose sequences. The model comprises N = 3 ASTGCN blocks with a latent representation size of 64 and a Chebyshev filter size of 7. Training is conducted over 100 epochs using MSE loss and the Adam optimizer with a learning rate of ${1e^4}$.

\begin{table}[h]
    \centering
    \resizebox{0.4\textwidth}{!}{
        \begin{tabular}{lcccccc}
            \toprule
            \textbf{Model} & \multicolumn{3}{c}{\textbf{mAP}} & \multicolumn{3}{c}{\textbf{Localization Latency (ms)}} \\
            \cmidrule(lr){2-4} \cmidrule(lr){5-7}
             & \textbf{Train} & \textbf{Test} & \textbf{Avg} & \textbf{Train} & \textbf{Test} & \textbf{Avg} \\
            \midrule
            DiveNet & -- & -- & -- & -- & -- & 23.65 \\
            STGCN & 71.66 & 74.89 & 73.27 & 51.67 & 48.17 & 49.92 \\
            TSAGCN & 75.91 & 73.96 & 74.93 & 38.34 & 50.00 & 44.17 \\
            AGCN & 80.00 & 80.36 & 80.18 & 32.67 & 32.17 & 32.42 \\
            \textbf{Ours} & \textbf{80.23} & \textbf{85.10} & \textbf{82.66} & \textbf{30.67} & \textbf{27.50} & \textbf{29.09} \\
            \bottomrule
        \end{tabular}
    }
    \caption{Comparison of mAP (in \%) and localization latency (in ms) during training and testing across different models.}
    \label{tab:map_latency_comparison}
\end{table}

To assess the efficiency of our proposed unsupervised pipeline, we evaluate the \textit{localization latency}~\cite{murthy2023divenet} and mean Average Precision (mAP). Table~\ref{tab:map_latency_comparison} compares our method against several baseline models, including supervised and self-supervised GNN-based approaches such as STGCN~\cite{yu2017spatio}, TSAGCN~\cite{shi2019two}, AGCN~\cite{li2018adaptive}, and the state-of-the-art DiveNet~\cite{murthy2023divenet}.

Our approach achieves an average localization latency of \textbf{29.09 ms}, with \textbf{30.67 ms} during training and \textbf{27.50 ms} during inference, outperforming all baseline GNN models. Furthermore, we also report a mean Average Precision (mAP) of \textbf{82.66\%} (80.23\% train, 85.10\% test), which is on par with or superior to fully supervised methods such as AGCN (80.18\% mAP) and significantly better than STGCN (73.27\%) and TSAGCN (74.93\%).

\subsection{Interpretability of ADM}
The Action Dynamics Metric (ADM) captures motion variations as a single-point representation over time. In diving sports, distinct ADM values correspond to sub-actions such as twists (rotation around a limb axis), somersaults (in-plane rotations), takeoff (forward or backward), pike, tuck, and water entry. Additionally, each diving sub-action exhibits a characteristic waveform with consistent critical points.

Figures~\ref{fig:fig2}(a) and (b) illustrate a somersault dive, where critical and inflection points highlight key phases. The initial takeoff is marked by the first critical point, with forward and backward takeoffs appearing as minima and maxima, respectively. Maxima (red) indicate the tuck position, where rotation is maximized, while minima (blue) denote the completion of one somersault and transition to the next. Inflection points signal action sequence changes. Conversely, Figure~\ref{fig:fig2}(c) depicts a twisting dive, where the ADM curve maintains an almost constant slope, reflecting minor action variations. The first inflection point corresponds to the pike position, initiating the twist, followed by minima and maxima representing minor subclass variations within the twisting dive.
\begin{figure*}[t!]
    \centering
    \includegraphics[width=0.9\textwidth]{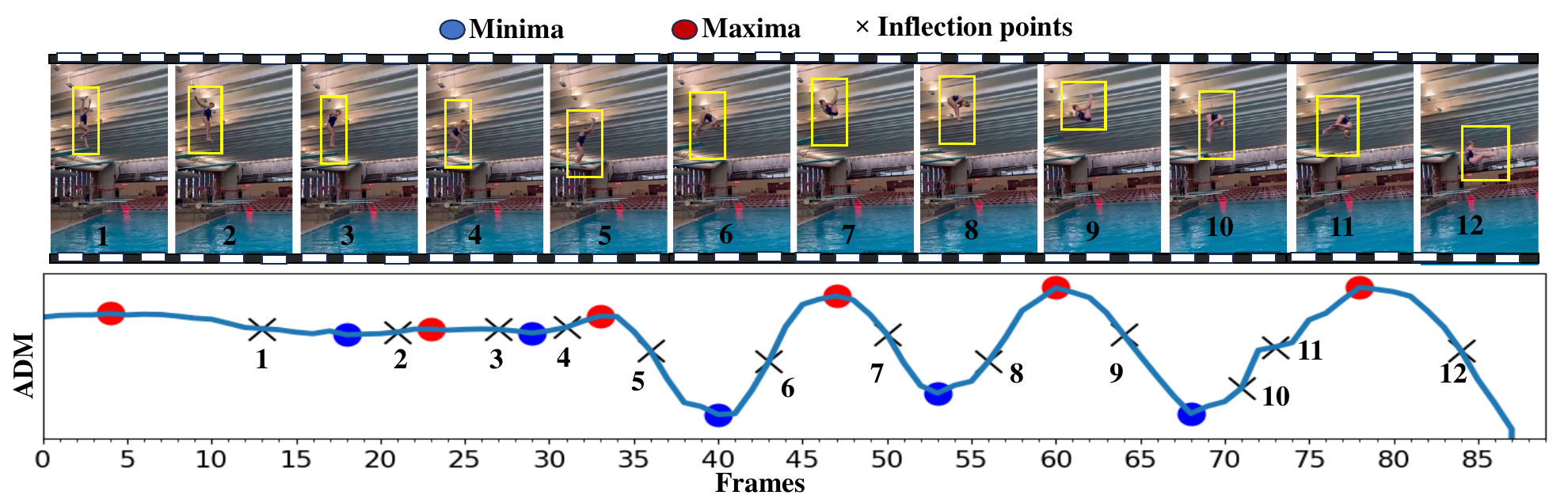}
    \caption{Variations in ADM curve for a YouTube diving video, highlighting critical and inflection points correspond to significant action transitions.}
    \label{demonstration}
\end{figure*}

\subsection{Ablation Study}
\label{sec:fig3}
We conducted ablation studies to analyze the impact of critical hyperparameters on model performance, focusing on embedding dimension, Chebyshev filter size, and the number of ASTGCN blocks. These experiments systematically fine-tune the model to optimize spatial-temporal feature extraction.


\begin{table}[ht]
    \centering
    \resizebox{0.44\textwidth}{!}{%
        \begin{tabular}{ccccc}
            \hline
            \textbf{No. of ASTGCN Blocks} & \textbf{EMD} & \multicolumn{3}{c}{\textbf{Localization Latency (ms)}} \\
            \cline{3-5}
            & & \textbf{Train} & \textbf{Test} & \textbf{Avg} \\
            \hline
            \multirow{3}{*}{2} & 64  & 64.5  & 61.67 & 63.09 \\
                                & 128 & 59.5  & 71.84 & 65.67 \\
                                & 256 & 64.5  & 56.5  & \textbf{60.5} \\
            \hline
            \multirow{3}{*}{3} & 64  & 47.5  & 54.67 & \textbf{51.09} \\
                                & 128 & 64.0  & 55.84 & 59.92 \\
                                & 256 & 54.0  & 65.17 & 59.59 \\
            \hline
            \multirow{3}{*}{5} & 64  & 62.34 & 70.67 & 66.51 \\
                                & 128 & 51.5  & 55.84 & 53.67 \\
                                & 256 & 44.67 & 57.67 & \textbf{51.17} \\
            \hline
            \multirow{3}{*}{7} & 64  & 53.67 & 76.17 & 64.92 \\
                                & 128 & 59.17 & 69.5  & 64.34 \\
                                & 256 & 63.34 & 58.67 & \textbf{61.01} \\
            \hline
        \end{tabular}%
    }
    \caption{Performance comparison with varying number of ASTGCN Blocks}
    \label{tab:table2}
\end{table}

The number of ASTGCN blocks (2, 3, 5, 7) and embedding dimensions (64, 128, 256) are varied in combination to identify the optimal configuration. As shown in Table~\ref{tab:table2}, an embedding dimension of 64 with three ASTGCN blocks achieved the best performance with a minimum localization latency of 56.09 ms. Subsequently, as listed in Table~\ref{tab:table3}, we fixed the embedding size at 64 and explored the Chebyshev filter sizes (3, 5, 7, 9, 11, 13), determining that a filter size of 7 yielded the lowest localization latency of 29.09 ms.


\begin{table}[h]
    \centering
    \resizebox{0.4\textwidth}{!}{%
        \begin{tabular}{lccc}
            \toprule
            \textbf{No. of Chebyshev Filters} & \multicolumn{3}{c}{\textbf{Localization Latency (ms)}} \\
            \cmidrule(lr){2-4}
            & \textbf{Train} & \textbf{Test} & \textbf{Avg} \\ 
            \midrule
            3  & 45.50 & 56.50 & 51.00 \\
            5  & 63.50 & 56.34 & 59.92 \\
            \textbf{7}  & \textbf{30.67} & \textbf{27.50} & \textbf{29.09} \\ 
            9  & 47.17 & 52.84 & 50.01 \\
            11 & 55.00 & 53.34 & 54.17 \\
            13 & 57.00 & 46.00 & 51.50 \\ 
            \bottomrule
        \end{tabular}%
    }
    \caption{Performance comparison with varying Chebyshev filter sizes.}
    \label{tab:table3}
\end{table}

\subsection{In-the-Wild Setting Testing}
\label{sec:print}

To illustrate the application, we present the performance of our technique on five diving practice session videos sourced from YouTube. We extract 2D pose sequences utilising the cutting-edge AlphaPose ~\cite{fang2022alphapose} and process them using our suggested workflow. Figure~\ref{demonstration} shows how the ADM identifies inflection points to record temporal action transitions between dive phases. These findings demonstrate the generalisability of our unsupervised learning system in recognising action transitions in unstructured motion data. 

\section{Conclusion}
In this work, we introduced an unsupervised deep learning framework for fine-grained action localization in diving videos, leveraging spatio-temporal graph embeddings to detect motion transitions. Our method eliminates the need for manual annotations by utilizing Action Dynamics Metric (ADM) to identify key action transition points. Experimental results on the DSV Diving dataset and unseen real-world videos validate the effectiveness of our approach, achieving competitive performance compared to supervised baselines. Despite its effectiveness, our approach relies on accurate 2D pose sequences, and errors in pose estimation can propagate through the pipeline, affecting localization accuracy. In future work, we will explore domain adaptation strategies to improve method transferability across sports without needing retraining. Additional ablation study and evaluation results can be found in the
\href{https://sigport.org/documents/utal-gnn-unsupervised-temporal-action-localization-using-graph-neural-networks}{Supplementary Material}.
\label{sec:refs}

\bibliographystyle{IEEE}
\bibliography{ref}

\begin{thebibliography}{10}

\bibitem{cardenas2024beyond}
Fernando~Pedro Cardenas~Hernandez, Jan Schneider, Daniele Di~Mitri, Ioana Jivet, and Hendrik Drachsler,
\newblock ``Beyond hard workout: A multimodal framework for personalised running training with immersive technologies,''
\newblock {\em British Journal of Educational Technology}, 2024.

\bibitem{li2020graph}
Jin Li, Xianglong Liu, Zhuofan Zong, Wanru Zhao, Mingyuan Zhang, and Jingkuan Song,
\newblock ``Graph attention based proposal 3d convnets for action detection,''
\newblock in {\em Proceedings of the AAAI Conference on Artificial Intelligence}, 2020, vol.~34, pp. 4626--4633.

\bibitem{guo2019attention}
Shengnan Guo, Youfang Lin, Ning Feng, Chao Song, and Huaiyu Wan,
\newblock ``Attention based spatial-temporal graph convolutional networks for traffic flow forecasting,''
\newblock in {\em Proceedings of the AAAI Conference on Artificial Intelligence}, 2019, vol.~33, pp. 922--929.

\bibitem{yang2020revisiting}
Le~Yang, Houwen Peng, Dingwen Zhang, Jianlong Fu, and Junwei Han,
\newblock ``Revisiting anchor mechanisms for temporal action localization,''
\newblock {\em IEEE Transactions on Image Processing}, vol. 29, pp. 8535--8548, 2020.

\bibitem{zhao2020bottom}
Peisen Zhao, Lingxi Xie, Chen Ju, Ya~Zhang, Yanfeng Wang, and Qi~Tian,
\newblock ``Bottom-up temporal action localization with mutual regularization,''
\newblock in {\em Computer Vision--ECCV 2020: 16th European Conference, Glasgow, UK, August 23--28, 2020, Proceedings, Part VIII 16}. Springer, 2020, pp. 539--555.

\bibitem{liu2021multi}
Xiaolong Liu, Yao Hu, Song Bai, Fei Ding, Xiang Bai, and Philip~HS Torr,
\newblock ``Multi-shot temporal event localization: a benchmark,''
\newblock in {\em Proceedings of the IEEE/CVF Conference on Computer Vision and Pattern Recognition}, 2021, pp. 12596--12606.

\bibitem{islam2021hybrid}
Ashraful Islam, Chengjiang Long, and Richard Radke,
\newblock ``A hybrid attention mechanism for weakly-supervised temporal action localization,''
\newblock in {\em Proceedings of the AAAI conference on artificial intelligence}, 2021, vol.~35, pp. 1637--1645.

\bibitem{lee2020background}
Pilhyeon Lee, Youngjung Uh, and Hyeran Byun,
\newblock ``Background suppression network for weakly-supervised temporal action localization,''
\newblock in {\em Proceedings of the AAAI conference on artificial intelligence}, 2020, vol.~34, pp. 11320--11327.

\bibitem{shi2020weakly}
Baifeng Shi, Qi~Dai, Yadong Mu, and Jingdong Wang,
\newblock ``Weakly-supervised action localization by generative attention modeling,''
\newblock in {\em Proceedings of the IEEE/CVF conference on computer vision and pattern recognition}, 2020, pp. 1009--1019.

\bibitem{min2020adversarial}
Kyle Min and Jason~J Corso,
\newblock ``Adversarial background-aware loss for weakly-supervised temporal activity localization,''
\newblock in {\em Computer Vision--ECCV 2020: 16th European Conference, Glasgow, UK, August 23--28, 2020, Proceedings, Part XIV 16}. Springer, 2020, pp. 283--299.

\bibitem{zhang2022automatic}
Yanting Zhang, Fuyu Tu, Zijian Wang, et~al.,
\newblock ``Automatic moving pose grading for golf swing in sports,''
\newblock in {\em 2022 IEEE International Conference on Image Processing (ICIP)}. IEEE, 2022, pp. 41--45.

\bibitem{fang2024bid}
Qihang Fang, Chengcheng Tang, Shugao Ma, and Yanchao Yang,
\newblock ``Bid: Boundary-interior decoding for unsupervised temporal action localization pre-trainin,''
\newblock {\em arXiv preprint arXiv:2403.07354}, 2024.

\bibitem{tang2023unsupervised}
Haoyu Tang, Han Jiang, Mingzhu Xu, Yupeng Hu, Jihua Zhu, and Liqiang Nie,
\newblock ``Unsupervised temporal action localization via self-paced incremental learning,''
\newblock {\em arXiv preprint arXiv:2312.07384}, 2023.

\bibitem{seweryn2023survey}
Karolina Seweryn, Anna Wr{\'o}blewska, and Szymon {\L}ukasik,
\newblock ``Survey of action recognition, spotting and spatio-temporal localization in soccer--current trends and research perspectives,''
\newblock {\em arXiv preprint arXiv:2309.12067}, 2023.

\bibitem{xu2022finediving}
Jinglin Xu, Yongming Rao, Xumin Yu, Guangyi Chen, Jie Zhou, and Jiwen Lu,
\newblock ``Finediving: A fine-grained dataset for procedure-aware action quality assessment,''
\newblock in {\em Proceedings of the IEEE/CVF conference on computer vision and pattern recognition}, 2022, pp. 2949--2958.

\bibitem{murthy2023divenet}
Pramod Murthy, Bertram Taetz, Arpit Lekhra, and Didier Stricker,
\newblock ``Divenet: Dive action localization and physical pose parameter extraction for high performance training,''
\newblock {\em IEEE Access}, vol. 11, pp. 37749--37767, 2023.

\bibitem{chen2020curvature}
He~Chen and Gregory~S Chirikjian,
\newblock ``Curvature: A signature for action recognition in video sequences,''
\newblock in {\em Proceedings of the IEEE/CVF Conference on Computer Vision and Pattern Recognition Workshops}, 2020, pp. 858--859.

\bibitem{yu2017spatio}
Bing Yu, Haoteng Yin, and Zhanxing Zhu,
\newblock ``Spatio-temporal graph convolutional networks: A deep learning framework for traffic forecasting,''
\newblock {\em arXiv preprint arXiv:1709.04875}, 2017.

\bibitem{shi2019two}
Lei Shi, Yifan Zhang, Jian Cheng, and Hanqing Lu,
\newblock ``Two-stream adaptive graph convolutional networks for skeleton-based action recognition,''
\newblock in {\em Proceedings of the IEEE/CVF conference on computer vision and pattern recognition}, 2019, pp. 12026--12035.

\bibitem{li2018adaptive}
Ruoyu Li, Sheng Wang, Feiyun Zhu, and Junzhou Huang,
\newblock ``Adaptive graph convolutional neural networks,''
\newblock in {\em Proceedings of the AAAI conference on artificial intelligence}, 2018, vol.~32.

\bibitem{fang2022alphapose}
Hao-Shu Fang, Jiefeng Li, Hongyang Tang, Chao Xu, Haoyi Zhu, Yuliang Xiu, Yong-Lu Li, and Cewu Lu,
\newblock ``Alphapose: Whole-body regional multi-person pose estimation and tracking in real-time,''
\newblock {\em IEEE Transactions on Pattern Analysis and Machine Intelligence}, vol. 45, no. 6, pp. 7157--7173, 2022.

\end{thebibliography}

\end{document}